\documentclass{article}
\usepackage{spconf,amsmath,graphicx}
\usepackage{booktabs}
\usepackage[pdftex]{hyperref}
\usepackage{multirow}
\usepackage{tabularx}
\usepackage{adjustbox}
\hypersetup{bookmarksopen,bookmarksnumbered,colorlinks=true,linkcolor=black,citecolor=blue,urlcolor=blue}

\usepackage{enumitem}
\setlist{nosep, leftmargin=14pt}

\usepackage{mwe} 

\title{{Spatio-temporal Transformer to support automatic sign language translation}}

\name{Christian Ruiz $^*$, Fabio Mart\'inez$^*$}
\vspace{-0.35cm}
\address{$^*$ Biomedical Imaging, Vision and Learning Laboratory (BIVL$^{2}$ab),\\ Universidad Industrial de Santander, Bucaramanga, Colombia\\}

\begin{document}
\maketitle
\begin{abstract}
 Sign Language Translation (SLT) systems support hearing-impaired people communication by finding equivalences between signed and spoken languages. This task is however challenging due to multiple sign variations, complexity in language and inherent richness of expressions.  Computational approaches have evidenced capabilities to support SLT. Nonetheless, these approaches remain limtied to cover gestures variability and support long sequence translations. This paper introduces a Transformer-based architecture that encodes spatio-temporal motion gestures, preserving both local and long-range spatial information through the use of multiple convolutional and attention mechanisms. The proposed approach was validated on the Colombian Sign Language Translation Dataset (CoL-SLTD) outperforming baseline approaches, and achieving a BLEU4 of 46.84\%.
 Additionally, the proposed approach was validated on the RWTH-PHOENIX-Weather-2014T (PHOENIX14T), achieving a BLEU4 score of 30.77\%, demonstrating its robustness and effectiveness in handling real-world variations.
\end{abstract}

%
\section{Introduction}
\label{sec:intro}

Sign language is the main alternative of communication for approximately 466 million hearing-impaired individuals worldwide \cite{who2018}. However, effective communication even among deaf people can be challenging due to the various cultural, regional, and socio-linguistic differences in sign language. Moreover, communication with the rest of society is severely limited due to the lack of knowledge of sign languages, creating a significant obstacle to full immersion of deaf people in daily activities. Hence, translation support alternatives are crucial to facilitate communication among hearing-impaired individuals and the rest of society \cite{woll2001multilingualism}.

Sign languages are complex visual languages that involve spatio-temporal parameters present in the articulation of signs such as motion, position, hand-shape and orientation as well as non-manual components corresponding to mouth and eyebrow shapes \cite{perniss2018mapping}. As a consequence, modeling, characterizing, and developing computational methods that can handle a continuous sign language translation remains a challenging task. \\

Currently, state-of-the-art approaches for sign language translation are mostly based on sequence-to-sequence models with encoder-decoder architectures \cite{camgoz2018neural}. These approaches are typically built on recurrent neural networks (RNN), which can limit the network's ability to identify long-range spatial dependencies inherent in sign language \cite{li2022sign, zhang2019sign}. To address these challenges, more advanced approaches such as Transformers have been proposed to capture long temporal dependencies in sign language translation, considering multiple attention mechanism to weight importance to key gestures in a communication \cite{camgoz2020sign}. Current SLT Transformers, however, lose spatio-temporal information given by sign gestures, as they typically use flattened input representations, which may result in the loss of significant information in sign communication. Also, during language communication, motion information is relevant to bring additional contextual information, and therefore, new schemes require the exploitation of spatio-temporal gestural pose estimation, as well as motion kinematic primitives.


The main contribution of this work is a Transformer-based architecture that takes in a more suitable spatial representation of sign gestures to encode both local and long range spatial dependencies. To achieve this, we enhance the encoding strategy by creating a better input representation which consists of first taking optical-flow images which helps the network focus on motion kinematic patterns and extract from these two-dimensional (2D) representations of convolutional feature maps in order to preserve relevant local spatial information present in gestures. We explore the application of 2D positional encodings to handle 2D data. Additionally, we include a 2D self-attention mechanism that takes the previously extracted feature maps and calculates a pixel-wise attention to better highlight long range dependencies while acting as a complement to convolutions. Our proposed approach has therefore the capability to detect variations related to both manual and non-manual articulators while trained using a sequence-to-sequence scheme. 



\section{Related Work}
\label{sec:current_work}

Sign language translation is the task of mapping a sign language video into its spoken language counterpart. Recently, computational approaches have been introduced to improve recognition of sign language instances in a continuous manner. Hao Zhou \textit{et al}., 2020. \cite{9354538} proposed an architecture that seeks to serve in the Continuous Sign Language Recognition (SLR) task. The main idea of this work is to identify the relevance of the parameters involved in the production of signs through a feature extractor process by employing a CNN as their backbone; then a temporal modelling next to a sequence learning process is performed in order to recognize a sequence of \textit{glosses} \footnote{Sign glosses are spoken language words that match the meaning of
signs.} from a sign language video. However, this approach focuses solely on the recognition of signs, putting aside the translation task. Kayo Yin \textit{et al}., 2020. \cite{yin2020better} explored the mapping between the sequence of glosses identified in \cite{9354538} and it's correspondence in spoken language as a text-to-text task. A Transformer-based architecture \cite{vaswani2017attention} is employed to first encode a proper representation of sign glosses that will be further decoded taking advantage of the Transformer's self-attention mechanisms to identify the temporal correspondences between both sequences. Nevertheless, in this approach the translation task heavily relies on the quality of the gloss-recognition process which also implies modularizing the translation while failing to fully exploit the potential of encoder-decoder-based architectures to perform the task in and end-to-end manner. On the other hand, Rodriguez Jefferson \textit{et al}., 2021. \cite{https://doi.org/10.1049/cvi2.12037} proposed an encoder-decoder architecture to tackle end-to-end sign language translation while exploring optical flow as a more suitable representation of sign features. Optical flow representations allow to highlight spacial kinematic patterns which are subsequently processed by Bidirectional Recurrent Neural Network (BRNN) units in conjunction with attention modules to better exploit more complex temporal relationships along video descriptors. This approach is mostly based on recurrence, which currently has been surpassed by the introduction of Transformer networks that have been proved to be more efficient regarding information processing while being more powerful identifying contextual relationships. On the same line, Camgoz Necati \textit{et al}., 2020. \cite{camgoz2020sign} proposed a Transformer network to perform sign language translation while introducing a sign language recognition system within the encoder. Although this approach has represented a significant advance in the introduction of continuous sign language translation systems in real-life scenarios; this work, as well as most of the aforementioned approaches, generate a 1D representation of sign language images that might lack of relevant spatial information associated to the dimentional downsampling carried out during the feature extractor process.\\

\begin{figure*}[ht!]
\centering
\includegraphics[scale=0.35]{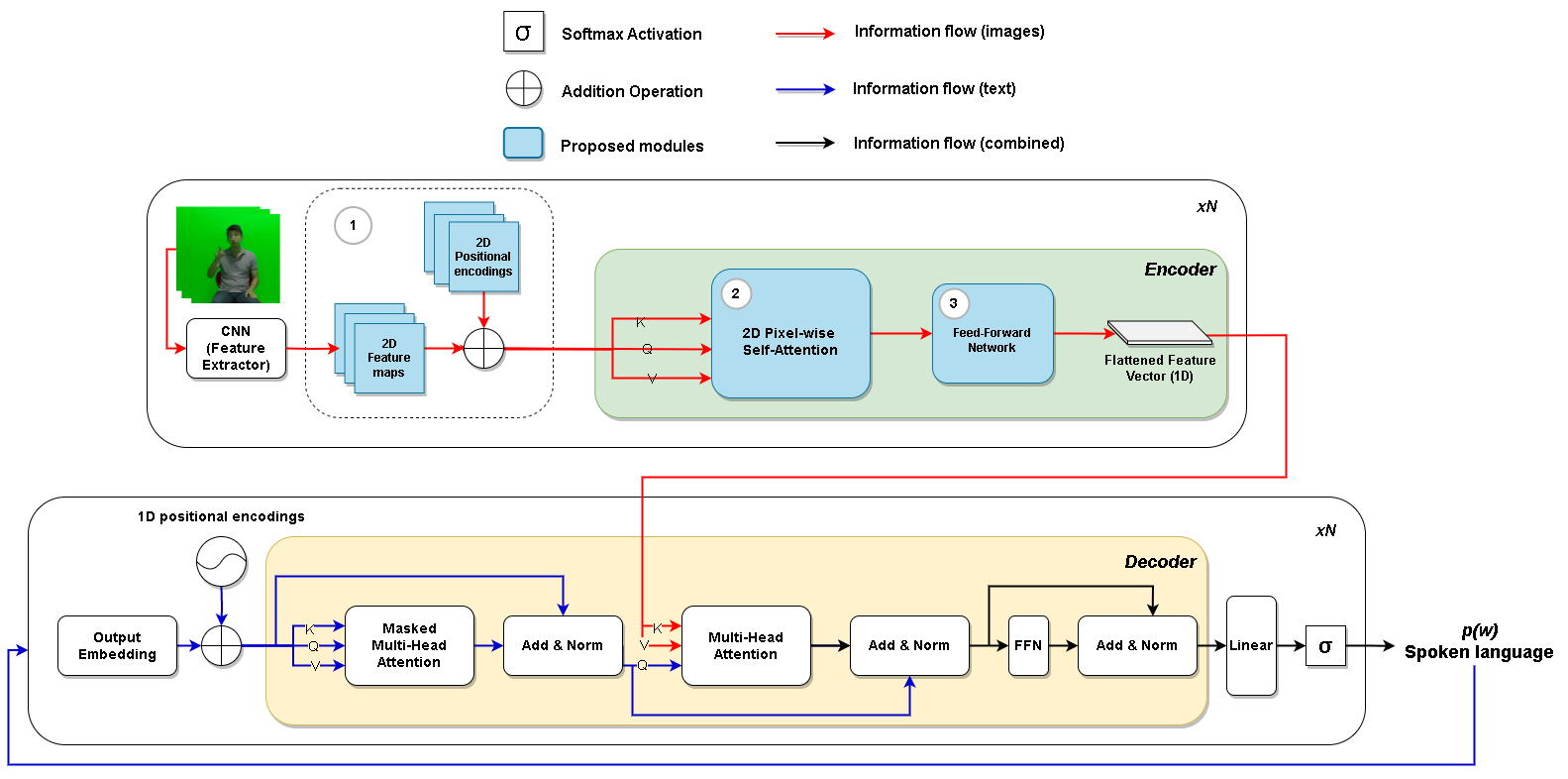}
\label{Fig:pipeline}
\caption{Pipeline of the proposed Transformer architecture. 1) Input representation. 2) Two-dimensional Self-Attention. 3) Feed-forward network.}
\end{figure*} 

\section{Proposed approach}
\label{sec:mat_and_met}

The proposed SLT network is summarized in Figure \ref{Fig:pipeline}. 

In this work, we aim to improve the encoding and utilization of spatial information in Sign Language Translation. To achieve this, we propose several key contributions. First, we focus on improving the input representation by retaining local spatial information. To achieve this, we convolve feature maps into a 2D matrix instead of a flattened feature vector. Additionaly, we explore two-dimensional positional encodings, which enhance the encoding of spatial information present in the feature maps. Second, we modify the Transformer's encoder by replacing their Multi-Head Attention (MHA) mechanism into a 2D pixel-wise self-attention mechanism that attends to long-range spatial dependencies. This allow us to identify the most significant parts of the image and effectively capturing the relationships between different parts of a gesture. Together, these contributions enable our model to encode and utilize both local and long-range spatial information more effectively, leading to more accurate translation of sign language.

\subsection{Input representation}

Convolutional neural networks (CNNs) are commonly used for image recognition tasks as they are able to automatically learn and extract relevant features from raw image data. We use a CNN backbone as a feature extractor for sign language images, convolving them into a 2D matrix instead of a flattened feature vector. This allows us to retain spatial locality information present in the image. In the traditional Transformer architecture, positional information is added to the embeddings using positional encodings, which makes the input invariant to permutations. We use a modified version of these sinusoidal positional encodings introduced in \cite{xu2021positional}  that is more suitable for two-dimensional input data. We use these positional encodings 2D as a way to retain the layout structure of the image which can carry critical semantics. We add each pixel a unique pair of coordinates (i, j), where i represents the row index and j represents the column index. The equations for the encodings are as follows:

\begin{align*}
\text{PE}(x,y,2i) &= \sin \left(\frac{x}{10000^{4i/D}}\right) \\
\text{PE}(x,y,2i+1) &= \cos \left(\frac{x}{10000^{4i/D}}\right) \\
\text{PE}(x,y,2j+\frac{D}{2}) &= \sin \left(\frac{y}{10000^{4j/D}}\right) \\
\text{PE}(x,y,2j+1+\frac{D}{2}) &= \cos \left(\frac{y}{10000^{4j/D}}\right)
\end{align*}

Here, (x,y) is a point in 2d space i,j is an integer in [0, D/4), where D is the size of the ch dimension. These positional encodings have the same size and dimension as the previously extracted feature maps and helps us capture the position of each pixel relative to other pixels in the image.

\subsection{2D Self-Attention}

Most Encoder-Decoder based models for SLT rely on convolutions as their feature extractor. However, convolutional layers process information within a local neighborhood, making it computationally inefficient to model long-range dependencies in images. We use a self-attention module that was previously proposed in the Generative Adversarial Network (GAN) framework \cite{zhang2019selfattention}. This module allows us to efficiently model relationships between spatial regions that are widely separated, which is crucial for distinguishing gestures that involve both manual and non-manual articulators. \\

Given a set of feature maps F with spatial dimensions $H \times W$ and $C$ channels, we can compute the self-attention map $A$ by first computing three sets of feature maps $Q$, $K$, and $V$. These feature maps are obtained by applying three different linear transformations to the original feature maps $F$:

\begin{align*}
Q &= W_Q F \\
K &= W_K F \\
V &= W_V F
\end{align*}

where $W_Q$, $W_K$, and $W_V$ are learned weight matrices.

Next, we compute the attention map $A$ as:

\begin{align*}
A &= \text{softmax}\left(\frac{QK^T}{\sqrt{d_k}}\right)
\end{align*}

where $d_k$ is the dimension of the key feature maps, and softmax is the softmax function that normalizes the attention weights. The attention map $A$ is a matrix with dimensions $H \times W \times H \times W$, and represents the importance of each spatial position in the feature maps.

Finally, we use the attention map $A$ to compute a weighted sum of the feature maps $V$:

\begin{align*}
O &= \text{softmax}\left(\frac{QK^T}{\sqrt{d_k}}\right)V
\end{align*}

where $O$ is the output of the self-attention mechanism, and represents a weighted sum of the original feature maps, where the attention weights have been applied. The output $O$ is then concatenated with the original feature maps $F$ and processed by a convolutional layer to produce the final output feature maps. The self-attention mechanism in SAGAN allows the network to focus on the most important parts of the input feature maps, leading to better generation of high-resolution images.

\subsection{Feed-forward Network}
Once we encoded our input representation through the 2D Self-Attention mechanism, we processed the output using a modified version of Transformers' feed-forward neural network, which consist of two convolutional layers. To improve stability and performance, we employed batch normalization, which prevents the outputs of each layer from drifting too far away from the input distribution and can prevent vanishing or exploding gradients during training. Finally, we turn the output into a flattened feature vector to create a more compact representation of the input sequence, which can be easily processed by the decoder.

\subsection{Decoder}

The decoder in our model uses the same architecture as the standard Transformer. Its main task is to take the output from the encoder, which consists of a sequence of hidden states representing an abstraction of the input sentence, and generate an output sequence in spoken language. The decoder is made up of three main components: the masked multi-head self-attention layer, the multi-head self-attention layer, and the feed-forward network. The masked multi-head self-attention layer helps the decoder focus on the current time step and ignore any future information to prevent overfitting, while the multi-head self-attention layer allows the decoder to attend to the most relevant parts of the input sequence, thereby generating more accurate output. Finally, the feed-forward network processes the output of the attention layers to generate the final output sequence. By using these three components, the decoder can effectively generate high-quality spoken output while considering context and spatio-temporal information.

\subsection{Data}

\subsubsection{CoL-SLTD}

The CoL-SLTD (Colombian Sign Language Translation Dataset) captures sign expressions using a conventional RGB camera, which facilitates natural sign language production. Each video sequence was recorded under controlled studio conditions with a green chroma key background, proper lighting, and the participant's position and clothing carefully controlled. The dataset consists of 39 sentences: 24 affirmative, 4 negative, and 11 interrogative. Each sentence has three different repetitions, resulting in a total of 1020 sentences, which allows for capturing sign motion variability associated with specific expressions. 

\subsubsection{PHOENIX-WEATHER-2014T}

The dataset is extracted from the weather forecast segments of the German TV station PHOENIX. It includes a parallel corpus of German sign language videos featuring 9 different signers. The videos have been annotated at the gloss-level with a vocabulary of 1,066 different signs, and translations into spoken German with a vocabulary of 2,887 different words. The dataset consists of 7,096 training pairs, 519 development pairs, and 642 test pairs.

\subsection{Experimental setup}

To validate our proposed network, we utilized sign language videos represented as a sequence of frames and converted them into optical flow representations. Each frame was passed through a pre-trained ResNet18 CNN architecture, which had been trained with ImageNet weights, to extract sign features from input images of size $[224\times224]$. We took the output before the pooling layer, which had dimensions of $[sequenceLength \times numberOfFeatures \times 7 \times 7]$, and added 2D positional encodings before feeding it into the Transformers' encoder, which was composed of a single layer. To implement the pixel-wise attention mechanism, we used a learnable parameter, Gamma, and included skipped connections to preserve local information extracted from the CNN. For the feed-forward network, we applied two convolutional layers with a kernel size of 3x3 and batch normalization. Gloss annotations were also used in the encoder with a Connectionist Temporal Classification (CTC) loss function to map each frame representation into a single gloss word. We used four heads for the standard multi-head attention mechanisms in the decoder. The network was trained for 10 epochs with a batch size of 1, using an Adam optimizer and a cross-entropy loss function.

\section{Evaluation and Results}

In this study, we investigated the effects of creating a more suitable spatio-temporal representation of sign language videos in order to retain relevant information that could lead to more accurate translations. We first evaluated the input representation, since our architecture works in an end-to-end manner, we took the sign language videos represented as a sequence of frames in RGB. We then evaluated our proposed approach on these images against our baselines as shown in Table \ref{tab:table1}

\begin{table}[h]
\centering
\footnotesize
\begin{tabular}{cccc}
\toprule
\multicolumn{1}{c}{Image Type (RGB)} & \multicolumn{1}{c}{RNN} & \multicolumn{1}{c}{Standard Transformer} & \multicolumn{1}{c}{P.A.} \\
\midrule
BLEU1 & - & 0.6730 & 0.6515 \\
BLEU2 & - & 0.5314 & 0.5421 \\
BLEU3 & - & 0.4327 & 0.4132 \\
BLEU4 & 0.1956 & 0.3277 & 0.3077 \\
\bottomrule
\end{tabular}
\caption{Comparison of Proposed Approach vs Baselines on RGB Images}
\label{tab:table1}
\end{table}

The results in Table \ref{tab:table1} indicate that both Transformer models outperformed the RNN approach on the reported BLEU4 score, which is expected since Transformers have demonstrated superiority in sequence-to-sequence tasks. However, our proposed approach did not achieve better performance than the Standard Transformer baseline when using RGB images. From this, we found that the Standard Transformer heavily relies on the order of tokens in the sequence due to its parallel processing, which makes it invariant to permutations. Although this is important for identifying sentence contexts in natural language processing, only relying on this limits its ability to incorporate spatial information effectively in the context of gesture recognition. To demonstrate this, we removed the one-dimensional positional encodings (1D P.E.) from the Standard Transformer baseline approach and evaluated its performance against our proposed approach (Table \ref{tab:table2}). We found that our proposed approach outperforms the baseline on all BLEU metrics when 1D P.E. are removed. This suggests that the Standard Transformer primarily learns the order of tokens rather than spatial information, which is key when identifying spatial variations in gestures.

\begin{table}[h]
\centering
\begin{tabular}{@{\extracolsep{\fill}}cccc}
\toprule
\multicolumn{1}{c}{Image Type (RGB)} & \multicolumn{1}{c}{Baseline - PE1D} & \multicolumn{1}{c}{P.A.} \\
\midrule
BLEU1 & 0.5730 & 0.6515 \\
BLEU2 & 0.4314 & 0.5421 \\
BLEU3 & 0.3327 & 0.4132 \\
BLEU4 & 0.2277 & 0.3077 \\
\bottomrule
\end{tabular}
\caption{Comparison of Proposed Approach vs Baseline without 1D P.E. on RGB Images}
\label{tab:table2}
\end{table}

To further investigate the effectiveness of our proposed approach, we also conducted experiments on a different type of sign language video representation. Specifically, we evaluated the performance of our approach on optical flow images. As shown in Table \ref{tab:table3}, we found that the proposed approach achieved significantly better results on optical flow images compared to RGB images. These findings suggest that optical flow images not only help identifying the gestures of the signer but also the movement over time which helps our network to fully exploit its capabilities.

\begin{table}[h]
\centering
\begin{tabular}{@{\extracolsep{\fill}}cccc}
\toprule
\multicolumn{1}{c}{} & \multicolumn{1}{c}{RGB} & \multicolumn{1}{c}{FLOW} \\
\midrule
BLEU1 & 0.5730 & 0.6515 \\
BLEU2 & 0.4314 & 0.5421 \\
BLEU3 & 0.3327 & 0.4132 \\
BLEU4 & 0.2277 & 0.3077 \\
\bottomrule
\end{tabular}
\caption{Comparison of Proposed Approach in both RGB and FLOW images}
\label{tab:table3}
\end{table}

For quantitative validation of the proposed approach, an ablation study was carried out to discover the contribution of each of the main components as well as to identify configuration strategies that support sign language translation. Table \ref{tab:table_ablation} summarizes the performance achieved by the proposed strategy when removing each of the proposed modules.

\begin{table}[h]
\centering
\small
\adjustbox{width=0.49\textwidth}{%
\begin{tabular}{|c|c|c|c|c|}
\hline 
 \textbf{METHOD} & \textbf{BLEU1} & \textbf{BLEU2} & \textbf{BLEU3} & \textbf{BLEU4} \\
\hline
 P.A. & 0.8235 & 0.7517 & 0.6283 & \textbf{0.5137} \\
 \hline
 P.A. - P.E 2D & 0.6905 & 0.6155 & 0.5417 & 0.4367 \\
\hline
 P.A. - ATTN 2D & 0.7130 & 0.6336 & 0.5027 & 0.4624 \\
\hline
 P.A. - FFN 2D & 0.7218 & 0.6415 & 0.5136 & 0.4620 \\
\hline
 P.A. - ATTN2D - FFN 2D  & 0.6515 & 0.5421 & 0.4132 & 0.2952 \\
\hline
 P.A. - GLOSSES & 0.7515 & 0.6421 & 0.5732 & 0.4699 \\
\hline
\end{tabular}
}
\caption{Proposed Approach - Ablation Study}
\label{tab:table_ablation}
\end{table}

From the table, it can be observed that the proposed approach (P.A.) achieves the highest BLEU4 score of 0.5137. However, when the two-dimensional positional encodings (P.E 2D) are removed, the performance drops significantly, and the BLEU4 score decreases to 0.4367. In contrast, removing the attention mechanism (ATTN 2D) or the feed-forward network (FFN 2D) results in only a slight decrease in performance. When both the attention mechanism and the feed-forward network are removed, the performance decreases significantly. On the other hand, when glosses are added to the proposed approach, the performance improves for all BLEU metrics. Therefore, it can be concluded that each of the main components of the proposed approach contributes significantly to the overall performance, and adding glosses further improves the performance. \\

\begin{figure}[h!tbp]
\centering
\includegraphics[scale=0.11]{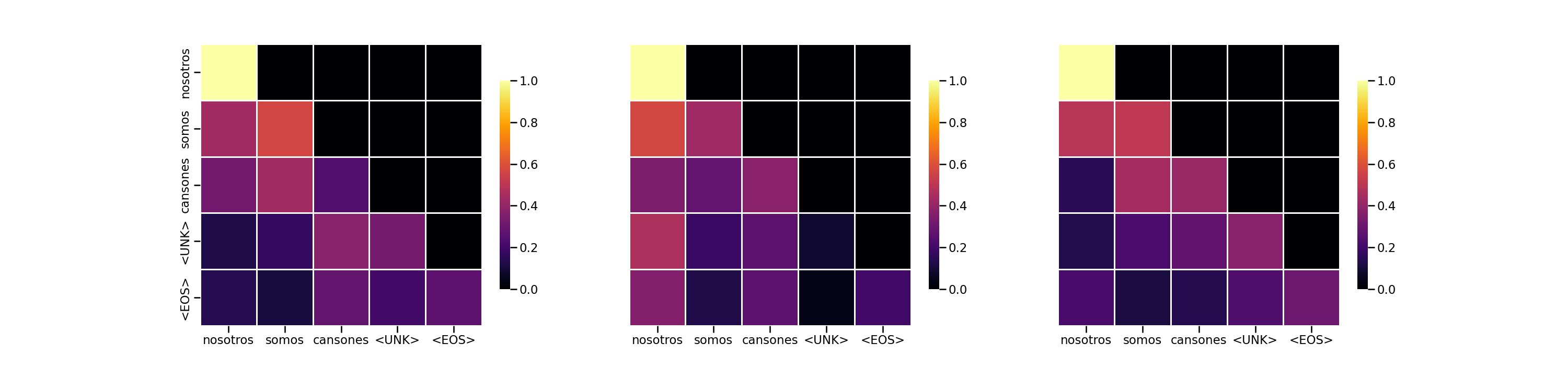}
\label{fig:results}
\caption{Attention maps from the self-attention mechanism in a Transformer's decoder. The input sequence $x_1, x_2, \ldots, x_n$ is passed through a self-attention mechanism, which produces attention maps that weight each element in the input sequence according to its relevance to each other element.}

\end{figure}

For qualitative results, we collected some of the most relevant visual outputs that could help to better understand the behavior of our approach. In Figure~\ref{fig:results}, we can see the attention maps from each head in the decoder's self-attention mechanism. From this, we can see how the networks identifies different relationships between each of the word of the sentence, thus allowing the network to learn relevant contexts.

In Figure~\ref{tab:predicted_real} on the other hand, we can see some of the text translations that are obtained compared to their ground truth counterpart. It is shown how even though the network could be quite accurate, there are still many cases where it fails to correctly translate the input sequence. These results suggest that further improvements can be made to the model to increase its overall performance.

\begin{table}[htbp]
  \centering
  \resizebox{0.5\textwidth}{!}{
  \begin{tabular}{ll}
    \toprule
    \textbf{Real sentence} & \textbf{Predicted sentence} \\
    \midrule
    nosotros somos felices \textless UNK\textgreater \textless EOS\textgreater & nosotros somos cansones que compró una casa \textless UNK\textgreater \textless EOS\textgreater \\
    carlos viaja a bogotá hoy \textless UNK\textgreater \textless EOS\textgreater & carlos viaja a bogotá hoy \textless UNK\textgreater \textless EOS\textgreater \\
    juan comprará un carro en el futuro \textless UNK\textgreater \textless EOS\textgreater & juan no comprará una casa en el futuro \textless UNK\textgreater \textless EOS\textgreater \\
   tu hermano tiene hambre \textless UNK\textgreater \textless EOS\textgreater & tu hermano tiene hambre \textless UNK\textgreater \textless EOS\textgreater \\
    a juan le gusta el chocolate \textless UNK\textgreater \textless EOS\textgreater & a juan le gusta esto y eso \textless UNK\textgreater \textless EOS\textgreater \\
    mary le cuenta a juan que compró una casa \textless UNK\textgreater \textless EOS\textgreater & mary le cuenta a juan que compró una casa \textless UNK\textgreater \textless EOS\textgreater \\
    \bottomrule
  \end{tabular}
  }
\caption{Predictions on LCSD Dataset}
\label{tab:predicted_real}
\end{table}

\newpage
\section{Conclusions}
\label{sec:conc}

This paper evaluated the effects of creating a more suitable spatio-temporal representation of sign language videos to improve accuracy in translation. The proposed approach was evaluated mainly on the Colombian Sign Language dataset (col-LSCD) against two baseline approaches that heavily relied on positional information regarding token position. It was found that even without this token-aware position information, our proposed approach [\ref{Fig:pipeline}] obtained similar results [\ref{tab:table1}] while only relying in attention and convolutional modules mainly focused on leveraging spatial information. The proposed approach was also evaluated on optical flow images and was found to achieve significantly better results [\ref{tab:table3}]. However, the proposed approach has some limitations. Specifically, it is computationally expensive, leading to long training times, and there might be some overfitting issues. Future work could focus on reducing the computational cost of the architecture to facilitate faster training times, especially when dealing with larger datasets. Input representation could be explored by implementing different convolutional architectures such as 3D convolutions, which can capture spatio-temporal features more effectively than 2D convolutions. The attention 2D module could be enhanced by introducing a multi-head-attention behaviour, which enables the model to attend to different parts of the input simultaneously. Finally, the feed-forward network can be further optimized by adding strategies such as depthwise convolutions as they can efficiently reduce the number of parameters needed in the model while preserving important spatial information. This can potentially improve the model's generalization ability and reduce the risk of overfitting, especially when dealing with limited training data. 
 
\section{Acknowledgments}
\label{sec:acknowledgments}

This work was partially funded by the Universidad Industrial de Santander. The authors acknowledge the Vicerrectoría de Investigación y Extensión (VIE) of the Universidad Industrial de Santander for supporting this research. They also gratefully acknowledge the support of NVIDIA Corporation with the donation of the Titan V GPU used for this research.

\bibliographystyle{IEEEbib}

\bibliography{refs}
\end{document}